\documentclass[a4paper,twoside]{article}
\usepackage{epsfig}
\usepackage{subcaption}
\usepackage{calc}
\usepackage{amssymb}
\usepackage{amstext}
\usepackage{amsmath}
\usepackage{amsthm}
\usepackage{multicol}
\usepackage{pslatex}
\usepackage{apalike}
\usepackage{graphicx} 
\usepackage{mathptmx} 
\usepackage{wrapfig}
\usepackage{microtype}
\usepackage{lipsum}
\usepackage{textcomp}
\usepackage{flushend}
\usepackage{SCITEPRESS}     

\begin{document}

\title{Evaluating Robot Posture Control and Balance by Comparison to Human Subjects using Human Likeness Measures}

\ifdefined\review
\author{\authorname{Authors hidden for double-blind review\sup{1}\orcidAuthor{0000-0000-0000-0000}}}
\else
\author{\authorname{Vittorio Lippi\sup{1}\orcidAuthor{0000-0001-5520-8974}, Christoph Maurer\sup{1}\orcidAuthor{0000-0001-9050-279X}, Thomas Mergner\sup{1}\orcidAuthor{0000-0001-7231-164X}}
\affiliation{\sup{1}University Clinic of Freiburg, Neurology, Freiburg, Germany}
\email{\{vittorio.lippi , christoph.maurer , thomas.mergner\}@uniklinik-freiburg.de}
}
\fi


\abstract{Posture control and balance are basic requirements for a humanoid robot performing motor tasks like walking and interacting with the environment. For this reason, posture control is one of the elements taken into account when evaluating the performance of humanoids. In this work, we describe and analyze a performance indicator based on the comparison between the body sway of a robot standing on a moving surface and the one of healthy subjects performing the same experiment. This approach is here oriented to the evaluation of human likeness. The measure is tested with three human-inspired humanoid posture control systems, the \textit{independent channel} (IC), the \textit{disturbance identification and compensation} (DEC), and the \textit{eigenmovement} (EM) control. The potential and the limitations connected with such human-inspired humanoid control mechanisms are then discussed.}

\onecolumn \maketitle \normalsize \setcounter{footnote}{0} \vfill
\section{INTRODUCTION}

The benchmarking of humanoid performance is gaining interest in the research community 
\ifdefined\review
\cite{conti2018people,torricelli2014benchmarking,torricelli2020benchmarking}.
\else
\cite{conti2018people,torricelli2014benchmarking,torricelli2020benchmarking,Lippi2020,Lippi2019}.
\fi
 The performance of a humanoid is a complex subject that covers several issues, e.g. sensor fusion, cognitive and motor aspects, mechanics, and energy efficiency. In particular, a recent European project, EUROBENCH, is proposing to implement standard and repeatable experimental procedures to evaluate and compare the performance of different robots. This work describes one specific posture control performance indicator implemented within the project, evaluating human-likeness based on the similarity between responses to external disturbances%
\ifdefined\review 
.
\else
, more are described in \cite{Lippi2019} and \cite{Lippi2020}.
\fi
Balance has been taken into account because falling is one of the typical reasons of 
failure for humanoids \cite{atkeson2015no,atkeson2018happened,guizzo2015hard}. In particular, the aspect of human likeliness is studied in this work. A formal and unanimous definition of human likeliness is still missing, although the concept is relevant both for robotics and neuroscience \cite{torricelli2014benchmarking}. 
\begin{figure}[t!]
	\centering
		\includegraphics[width=1.00\columnwidth]{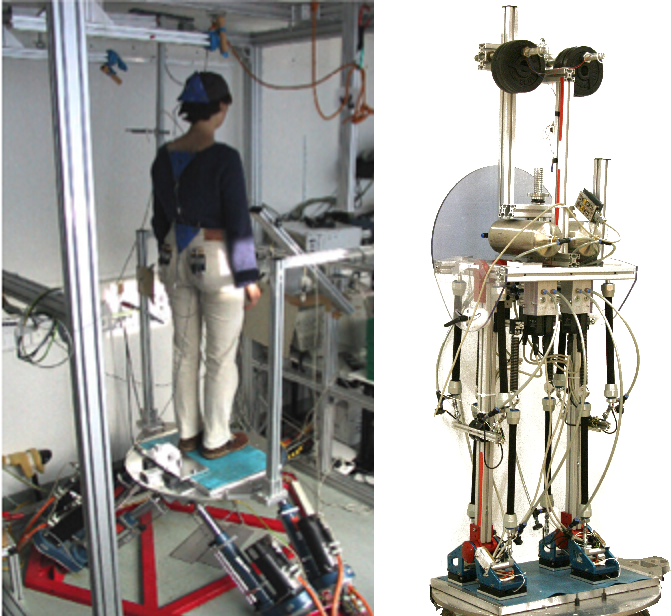}
	\caption{Posture control experiment with a moving support surface, a 6-Dof Stuart Platform. On the left, a human subject with active markers visible on the back (blue triangular plates) is shown. On the right, a robot standing on the same platform.}
	\label{fig:humanRobot}
\end{figure}
Human-like behavior is expected to entail some desirable features such as mechanical compliance (that is important for safety) and low energy consumption. Both of these features are reasonably associated with the low feedback loop gain imposed on biological systems in face of the presence of neural delay. In robotics the delay associated with the transport of signals is negligible, but the robustness in face of delays may be important in limiting the required computational resources needed to implement the real-time control \cite{ott2016good}. Although being often considered important, human-likeness at the state of the art is defined in terms of perception from the point of view of human observers \cite{von2013quantifying,oztop2005human,abel2020gender} or identified with the presence of a specific feature, e.g. reproducing human trajectories \cite{kim2009stable} or exhibiting compliance \cite{colasanto2015bio}. The measure proposed here is based on body kinematics to make the evaluation of human likeness repeatable and objective instead of designed \textit{ad hoc} and subjective.
Specifically, the evaluation is based on a data-set of results from human experiments. The experimental data consist of the body sway of human body segments induced by an external stimulus, here specifically the tilt of the support surface in the body sagittal plane. The measured body sway is characterized as a frequency response function, FRF, i.e. an empirical transfer function between the PRTS stimulus and the response computed on specific frequency points. The data-set has been produced with healthy subjects. The comparison aims to assess the similarity in the balancing behavior between the robot and an average healthy subject. Such measure does not make explicit assumptions about specific advantages of the human behavior, in contrast to the performance indexes proposed in other works that identify some goal like minimum torque or energy consumption in robots \cite{icinco16}, or a specific problem in human subjects \cite{singh2016detection}. The implications of the proposed measure and the limitations connected with human likeliness will be discussed in the conclusions.
\section{MATERIALS AND METHODS}
Comparing body sway profiles FRFs has been chosen as a base for the benchmarking because such analysis has been studied in several publications with human subjects. This provides a reference for comparison and tools for analysis. There are several reasons why the body sway induced by support surface tilt has been repeatedly used to study human posture control. As a repeatable stimulus it can be used to formally characterize the behavior in terms of responses to a well defined input. Furthermore, the tilting support surface requires the balancing mechanism to integrate vestibular input and proprioception (and vision, with eyes open), and hence it is well apt to study human sensor fusion. \\
\begin{figure}[htbp]
		\includegraphics[width=0.88\columnwidth]{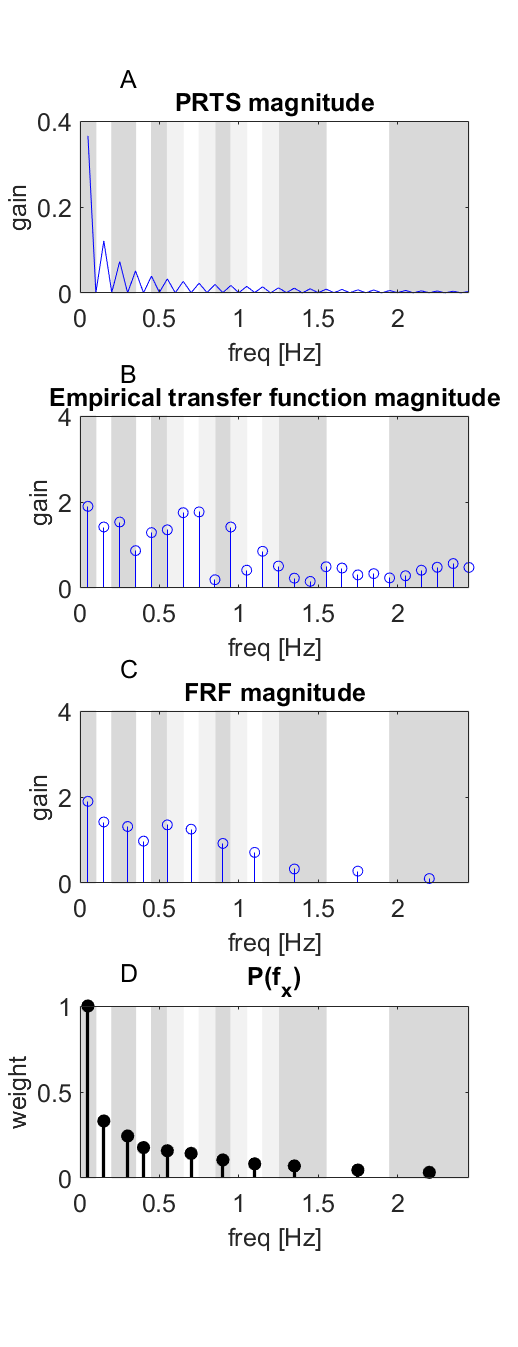}
	\caption{PRTS spectrum and averaging process. The magnitude of the discrete Fourier transform of the PRTS is shown in \textbf{A}. An empirical transfer function from a trial is shown in \textbf{B}, and the associated 11-component FRF are shown in \textbf{C}. Note that, as the FRF is averaged in the complex domain the magnitude of the average shown in the plot is not the average of the magnitudes. The bands on the background show the frequency ranges over which the spectrum is averaged: white and dark grey represent ranges associated with groups of frequencies. The sets of frequencies are overlapping, with light gray bands belonging to both the two contiguous groups, and a sample on the transition between two bands belongs to both the groups. The weights in \textbf{D} represents the values of $P(f_x)$ in eq. \ref{distance}.} 
	\label{fig:PRTSexample}
\end{figure}
\textbf{The sample data-set} here considered consists of data from human posture control experiments. A group of 38 young healthy subjects serves as a reference for the human-likeness criterion. The raw data include information like age, body weight, height, and the height of the different markers. In the experiments, the subjects were presented with a stimulus consisting of a tilt of the support surface in the body sagittal plane, while the recorded output was the body sway. The typical set-up is shown in Fig. \ref{fig:humanRobot}, the tracking was performed using active markers (Optotrak 3020; Waterloo, ON, Canada), attached to subjects' hips and shoulders and to the platform. A PC with custom-made programs was used to generate the support surface tilt  in the body-sagittal plane. The marker positions were recorded at 100 Hz using software written in LabView (National Instruments; Austin, TX, United States). The profile used for the stimulus is a pseudorandom ternary signal, PRTS \cite{peterka2002sensorimotor}. The peak-to-peak amplitude was set to $1^{\circ}$. The amplitude is rather small compared to what a healthy subject can withstand. Usually in similar studies the tilt may go up to 8 degrees and more \cite{peterka2002sensorimotor,asslander2015visual}. This was motivated by the aim to provide a data-set that could be compared safely with elderly subjects and patients and, in the specific case of robotic benchmarking, can be used to characterize the behavior of the robot without the risk of making the robot fall (and potentially break). \label{dataset}%
\\	\textbf{The performance indicator} is here a measure of similarity with human behavior. The body sway profiles, i.e. the angular sway of the body COM with respect to the ankle joint, is used to characterize and compare the responses. The comparison is defined in terms of the norm of the difference between frequency response functions (FRFs). The PRTS power-spectrum has a profile with peaks at  $f_{peak}$ separated by ranges of frequencies with no power \cite{jilk2014contribution,peterka2002sensorimotor,icinco20}. %
The response is considered for a specific set of frequencies where the PRTS spectrum has peaks: $\mathbf{f_{peak}}$=[0.05, 0.15, 0.25, 0.35, 0.45, 0.55, 0.65, 0.75, 0.85, 0.95, 1.05, 1.15, 1.25, 1.35, 1.45, 1.55, 1.65, 1.75, 1.85, 1.95, 2.05, 2.15, 2.25, 2.35, 2.45] Hz. Such a discrete spectrum is then transformed into a vector of 11 components by averaging the FRF over neighboring frequencies as illustrated in Fig. \ref{fig:PRTSexample}:
	
	\begin{equation}
		f_{x(k)}=\frac{\sum_{i \in B_k} f_{peak(i)}}{N_k}
		\label{bands}
	\end{equation}
	where $k$ and $i$ are the indexes of the components of the frequency vectors, $B_k$ is a set of $N_k$ frequencies averaged to obtain the $k^{th}$ sample. The $B_k$ are shown in Fig. \ref{fig:PRTSexample} as white and gray bands (notice that the bands are overlapping). Similarly the Fourier transform of the PRTS $P(f_{x})$ and the Fourier transform of the responses are averaged over the bands $B_k$ before computing the FRFs.   
The final representation of the FRF is a function of the 11 frequencies $f_{x}$ =[0.1, 0.3, 0.6, 0.8, 1.1, 1.4, 1.8, 2.2, 2.7, 3.5, 4.4]%
\ifdefined\review 
.
\else
\footnote{In a previous description of the system in \cite{Lippi2020} the frequency vector had 16 components as proposed in other works using PRTS, e.g. \cite{peterka2002sensorimotor,goodworth2018identifying,icinco20,lippi2020body}. In this work, we decided for a shorter version of the signal which is considered easier for human subjects (sometimes with elderly patients) and convenient for robotics experimenters. The discrete Fourier transform of the signal was consequently shorter, as well as the resulting FRFs.}.
\fi
The choice of the frequencies in $B_k$ and their overlapping follows the method described in \cite{peterka2002sensorimotor} but here is adapted to a vector of 11 frequencies. The rationale behind such a choice was to get a representation with the frequencies equally spaced on logarithmic scale, which is often used in posturography papers to present the FRF, with the overlapping providing a smoothing effect. 
In detail, the FRF is computed from the 11 components of the Fourier transform of the input $U$, and the output $Y$ as
\begin{equation}
	FRF=\frac{G_{UY}}{G_U}
\end{equation}
where $G_{UY}=U^* \odot Y$ and $G_U = U^* \odot U$ are empirical estimations of the cross power spectrum and the input power spectrum (``$\odot$'' is the  Hadamard, element-wise product). 
The peaks of the PRTS power-spectrum have larger values at lower frequencies \cite{jilk2014contribution}. This implies a better signal-to-noise ratio for the first components. A weighting vector $\mathbf{w}$ based on $P(f_{x})$ then defined in a similar way to eq. \ref{bands}, but considering the power 
\begin{equation}
	w_k=\sqrt{\sum_{i \in B_k} ||P(f_{peak(i)})||^2}
\end{equation}
The distance between two FRFs is defined and the norm of the difference weighted by the precision matrix, i.e. the inverse of the covariance matrix $\Sigma$, computed for the data-set of normal subjects. Before doing this the FRF is expanded into a vector with the real and imaginary components as separated elements, i.e. 22 components. This together with the foretold weighting leads to the definition of the norm:
\begin{equation}
D=\sqrt{S \Delta^T \Sigma^{-1} \Delta S}
\label{distance}
\end{equation}
where $S=diag([\mathbf{w},\mathbf{w}])$ is the diagonal matrix representing the re-weighting due to the power-spectrum, doubled to cover the 22 elements, and $\Delta$ is the difference between the two FRFs expanded to 22 components.
This approach does not require model identification because it is performed on the basis of the data. The comparison can be performed between the tested robot and the average of the groups of humans (healthy or with special deficient conditions) or between two single samples to quantify how much two robots differ from each other. The score of human-likeness is obtained comparing the sample with the mean of the human sample set $\mathbf{\mu}$ so that $\Delta=FRF-\mathbf{\mu}$. The parameters $\mathbf{\mu}$ and $\Sigma$ defining the score are shown in Fig. \ref{fig:params}.
\begin{figure}[tbp]
	\centering
		\includegraphics[width=1.00\columnwidth]{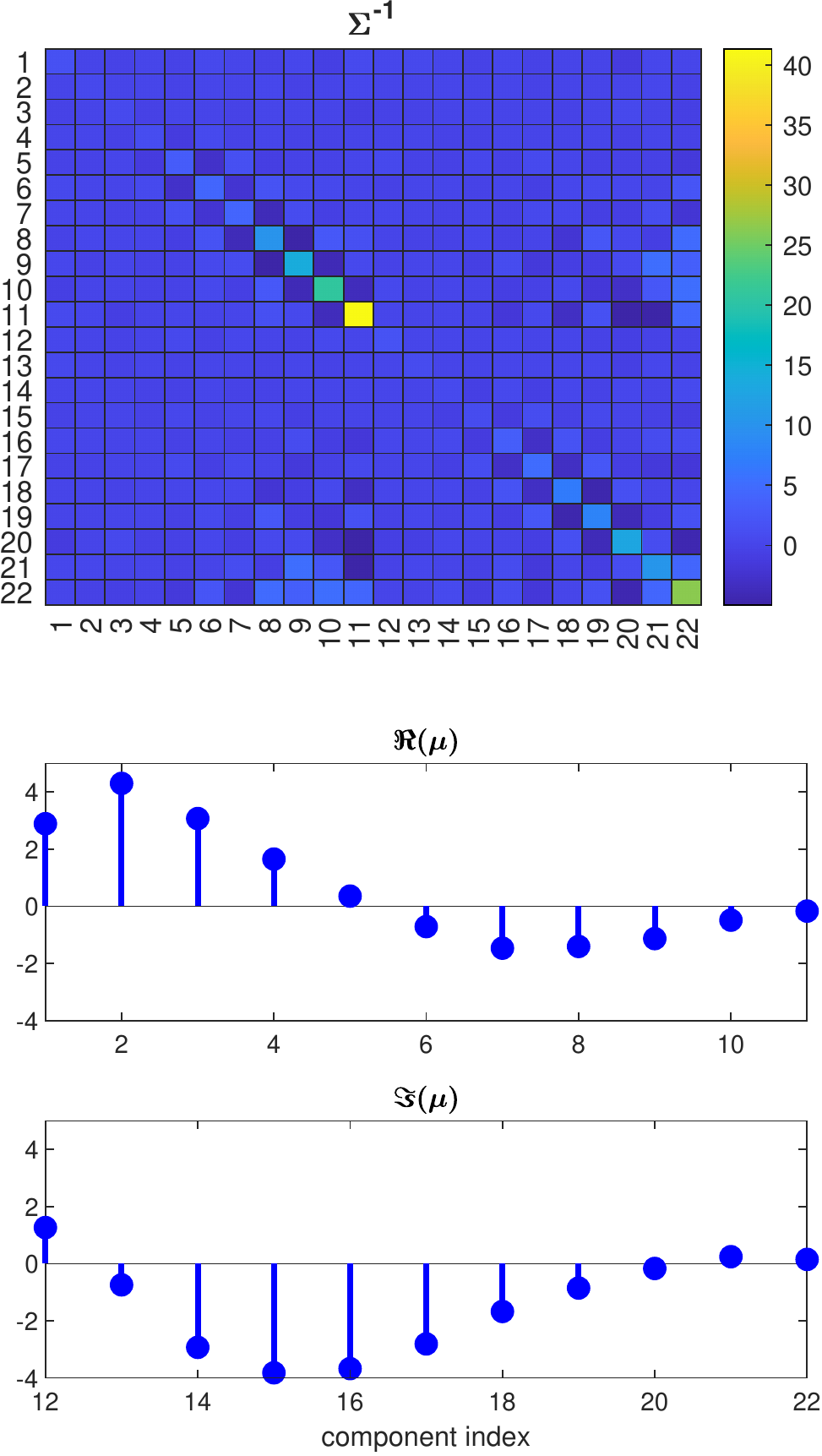}
	\caption{Parameters $\Sigma$ (top) and $\mathbf{\mu}$ (below) defining the score. Notice how the values on the diagonal corresponding to the highest frequencies are associated with high values (less variance in the dataset). This would make the metrics particularly sensitive to accidental changes in such components (e.g. due to noise). This disadvantage is compensated by the weighting profiles in Fig. \ref{fig:PRTSexample} D, which associate almost zero weight to high-frequency components.}
	\label{fig:params}
\end{figure}%
\begin{figure*}[t!]
	\centering
		\includegraphics[width=1.00\textwidth]{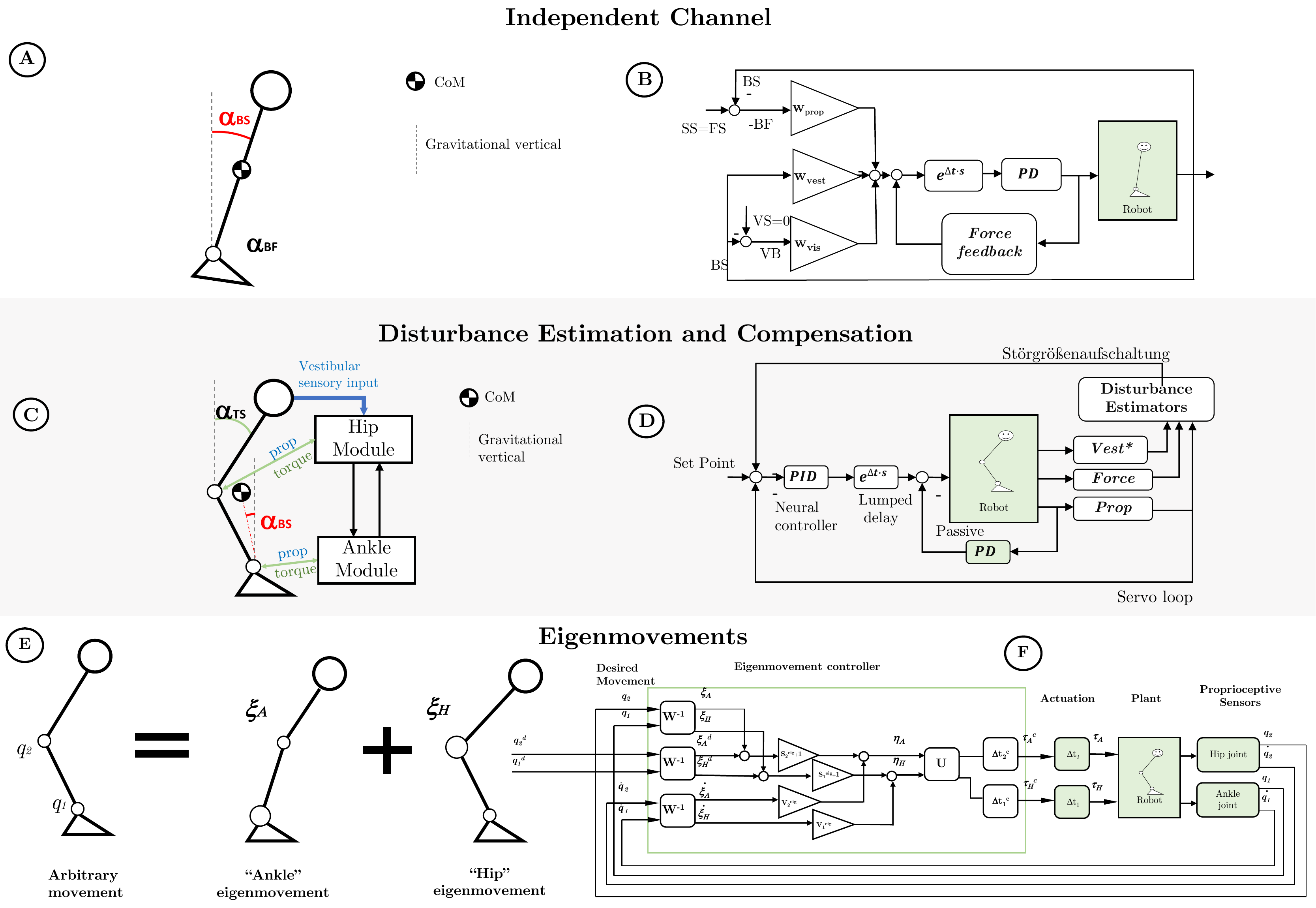}
	\caption{Overview of the tested three control systems. The notation is slightly different to match the one proposed by the respective authors in previous works. \textcircled{\tiny A} The single inverted pendulum mode for the IC, characterized by the body orientation in space $\alpha_{BS}$ and the angle between body and foot $\alpha_{BF}$.  \textcircled{\tiny B}. The IC control scheme as presented in \cite{10.3389/fncom.2018.00013}. The three weights $W_{prop}$, $W_{vest}$, and $W_{vis}$ are associated with proprioception ($\alpha_{BF}$), vestibular signal ($\alpha_{BS}$) and vision ($\alpha_{BS}$) respectively. The lumped delay $e^{\Delta t \cdot s}$ represents all the motor and sensory delays in the loop. The positive forcre feedback includes a gain and a first-order low pass filter: $K_F / (T_f s + 1)$. It is included to explain the observed behavior at low frequencies \cite{peterka2003simplifying}. \textcircled{\tiny C} The DEC, implemented as double inverted pendulum (DIP) \cite{G.Hettich2014,Hettich2013,hettich2015human}, with two control modules: the ``Hip Module'' controlling the orientation of the trunk in space $\alpha_{TS}$, and the ``Ankle Module'' controlling the orientation of the body CoM in space $\alpha_{BS}$. \textcircled{\tiny D} The scheme of a module of the DEC control. The system includes a passive control loop (PD) and a neural controller (PID) implementing the servo loop and the compensation of estimated disturbances as a \textit{st{\"o}rgr{\"o}{\ss}enaufschaltung}, i.e. feedforward compensation based on sensory input. There is a lumped delay $e^{\Delta t \cdot s}$ in each of the modules. The vestibular input \textit{vest*} for the ankle module uses a sensor fusion derived information (vestibular + proprioceptive) to reconstruct the orientation of legs in space. The vestibular input provides the orientation of the head in space (and hence of trunk in space) directly.  
 \textcircled{\tiny E} The decomposition of an arbitrary movement ($q_1$ = ankle angle, $q_2$ = hip angle) in two EMs.\textcircled{\tiny F} The EM control scheme.}
	\label{fig:ControlSystemsCrop}
\end{figure*}
\\ \textbf{Three control systems} are tested: the \textit{independent channel model} (IC), the \textit{disturbance estimation and compensation} (DEC), and the \textit{eigenmovements} concept (EM). Next we provide a brief description of the concepts.
\\ 1) The IC model is a simplified descriptive linear model for human upright posture control. The model consists of a single inverted pendulum controlled by a feedback mechanism with a PD controller and a time delay. The sensory integration mechanism consists of a weighted sum of the sensory contributions. The weights are constrained to unity \cite{peterka2002sensorimotor}. The three channels are: angle between body and foot perceived through proprioception, body-in-space orientation perceived by the vestibular system, and body-in-space orientation perceived by vision. The time delay represents all the delays in the control loop. In order to simulate a robot control experiment similar to the ones performed with the robot Posturob II in \cite{10.3389/fncom.2018.00013} the system has been simulated in ``closed-eyes'' conditions. The coefficient $W_{vis}$ is thus set to zero. 
\\ 2) The DEC model	was proposed as a model of human posture control in steady state \cite{Mergner2003}. It was implemented into a humanoid control system \cite{lippi2017human,lippi2016human} and tested on robotic platforms \cite{G.Hettich2014,lippi2018prediction,ott2016good,zebenay2015human}. The DEC exploits sensor-fusion-derived internal reconstructions of the external disturbances having impact on body posture and balance. The reconstructed disturbances are compensated using a servo controller. The model is nonlinear because it includes sensory thresholds and the disturbance estimators are non-linear functions. The control loop includes a lumped delay (one for each joint). Its formulation leads to a modular control system with one control module for each degree of freedom as shown in Fig. \ref{fig:ControlSystemsCrop} (C).
\\ 3) The EM \cite{alexandrov2005feedback} as implemented in \cite{alexandrov2017human}. This control system of bio-mechanical inspiration is based on the diagonalization of the inertial matrix $B_0$ describing the linearized body dynamics, i.e. 
\begin{equation}
B_0 \ddot{\mathbf{q}}-G_0\mathbf{q}=\tau_{con} 
\end{equation}
where $\mathbf{q}$ is the vector of joint angles, $G_0$ is the gravity matrix, and $\tau_{con}$ the control torques applied to the joints. EMs have independent dynamics, i.e. PD controller is implemented for each EM. In the original formulation the body kinematics is expressed in the joint angles $q_1$ and $q_2$ of Fig. \ref{fig:ControlSystemsCrop} E. The control of balance in space is here implemented by integrating a vestibular system (technically an IMU, \textit{inertial measurement unit}) and defining $q_1$ as leg-in-space sway angle (i.e. the sum of the hip angle and vestibular trunk in space orientation). When simulating the human system, there are different neural delays for the ankle and trunk control loops. The additional delays in Fig. \ref{fig:ControlSystemsCrop} F are added to make the delays the same for both the joints, whic here is assumed as a condition for decoupling the dynamics of the EM.  
\begin{figure}[htbp]
	\centering
		\includegraphics[width=1.00\columnwidth]{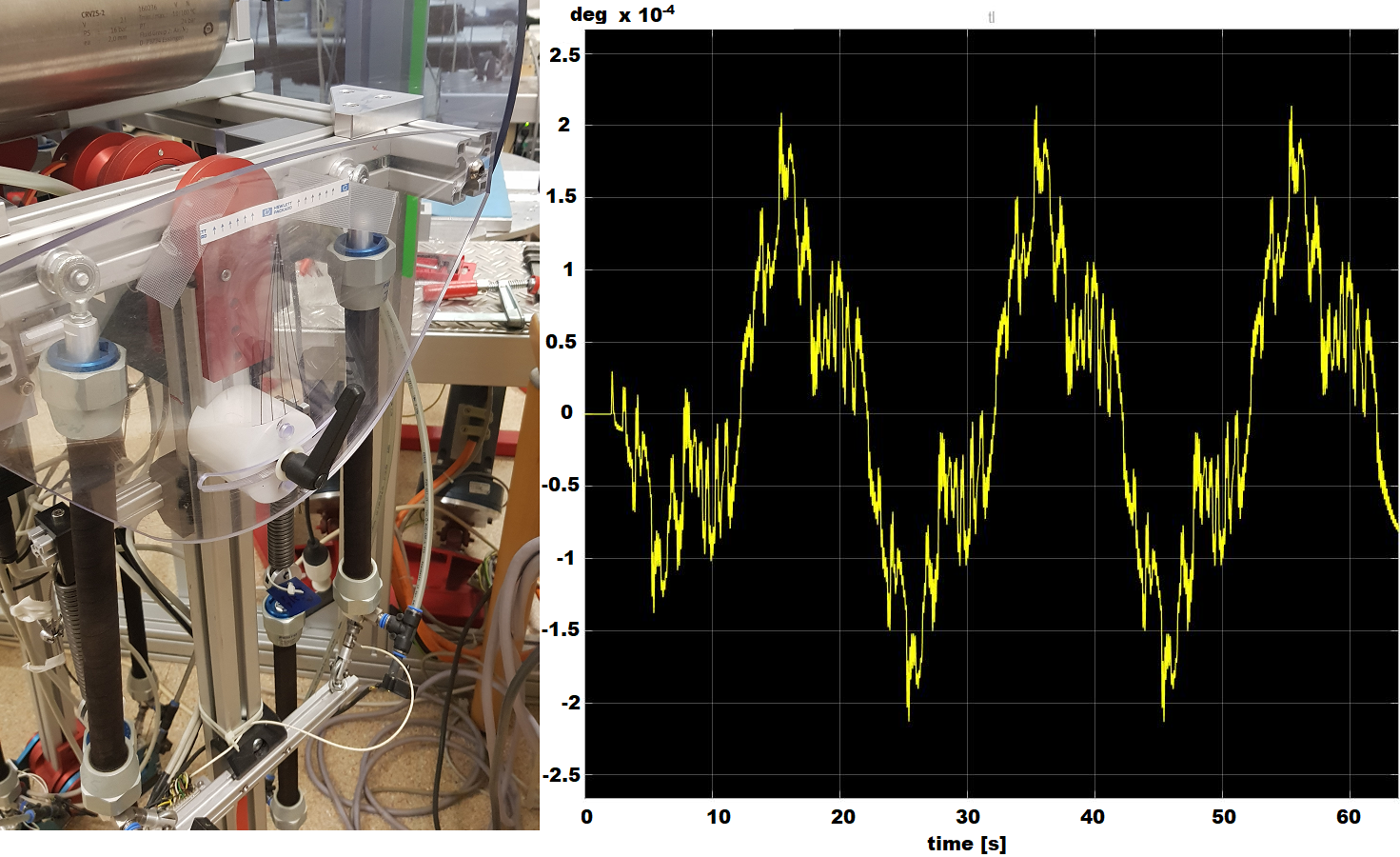}
	\caption{The hip joint of the robot Posturob II blocked by a screw (left) and the trunk to leg sway over time in simulation in degrees (right).}
	\label{fig:TLblocked}
\end{figure}

The parameters for the simulations are based on the experiments previously performed with the robot Posturob II using an IC, DEC, and EM approach as presented in \cite{10.3389/fncom.2018.00013}, \cite{G.Hettich2014}, and  \cite{alexandrov2017human}, respectively. The experiments are simulated in MATLAB/Simulink\textregistered. The robot is simulated using the double inverted model from \cite{albakri2008development}. In order to implement the SIP control for the IC, the hip stiffness is set to a large value, $\approx 10^5$ Nm, leading to virtually no movement ($\leq 10^{-3}$ deg, as shown in Fig. \ref{fig:TLblocked}). This is similar to what was implemented in \cite{10.3389/fncom.2018.00013} blocking the hip joint of the robot with a screw. This choice of simulation parameters aims to reproduce the robot behavior. The parameters of the models can be optimized to reproduce the human FRF our in simulation as previously done in posture control studies, e.g. \cite{asslander2015visual}. While this is useful to discuss the parameters themselves, e.g. to estimate the relative weight of different sensory channels in the control loop in different conditions, it is not obviously leading to a human-like behavior when applied on real robots. The responses are compared with eyes-closed human experiments and hence the visual signal is not considered: a trunk-in-space orientation is assumed to be available in the simulations with no specific modeling  of the dynamics of the vestibular system (which is, in general, faster but less accurate than the visual signal). This is because, at the state of the art, humanoids are not (yet) using visual input for balance control. This may change in the future considering the current advancements in computer vision and their integration in humanoid robotics. Data-sets for the eyes-open condition are however available for future benchmarking. 

\section{RESULTS}
\begin{figure}[htbp]
	\centering
		\includegraphics[width=1.00\columnwidth]{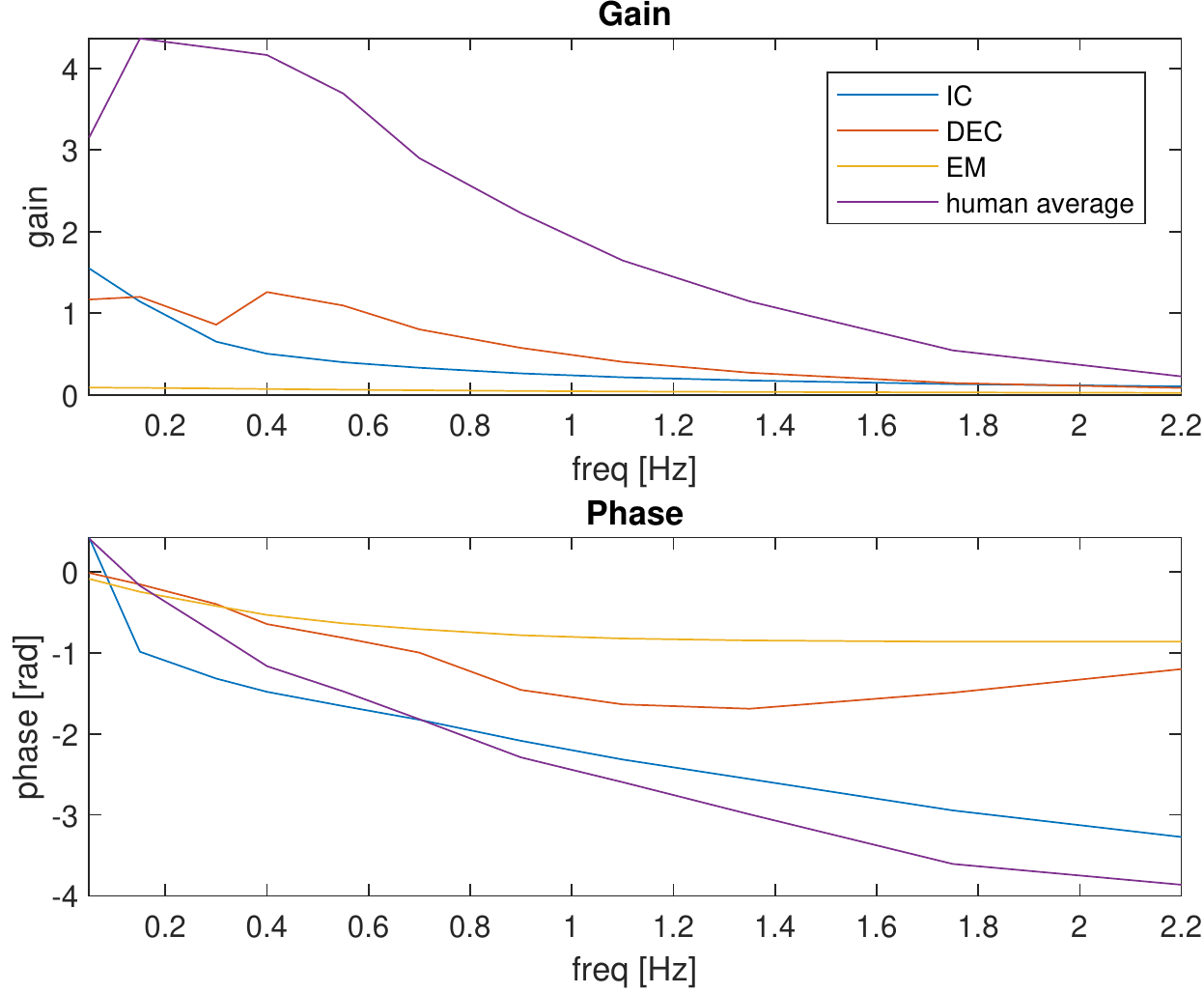}
	\caption{FRF obtained with the IC, DEC, and EM as well as the average FRF of the human data-set.}
	\label{fig:GainPhaseResponses}
\end{figure}

The simulation with the three control systems produced the FRFs shown in Fig. \ref{fig:GainPhaseResponses} together with the average of the human FRFs in the reference data-set. The results obtained were:
\begin{table}[h!]
\begin{center}
\begin{tabular}{lllll}
\textbf{control} & \textbf{score} &  \textbf{CDF}  \\ \cline{1-3}
\textit{IC}      & 2.2034         &  75.2212 \%  \\
\textit{DEC}     & 2.7318         &  87.6106 \%  \\
\textit{EM}      & 3.9616         &  97.3451 \%  \\
\end{tabular}
\end{center}
\end{table}
 The values in the third column give estimates of the cumulative distribution function (CDF) of the data-set at the given score. It is computed by counting the fraction of samples with a score smaller than the one produced in the specific simulation. The comparison in Fig. \ref{fig:GainPhaseResponses} shows that all three robotic controllers exhibit a smaller gain than the one observed on average in humans (especially the EM) and differences of phase. In Fig. \ref{fig:dist} the result for the three controllers is compared with the distribution of the human data-set.

\section{DISCUSSION, CONCLUSIONS AND FUTURE WORK}
\begin{figure}[t]
	\centering
		\includegraphics[width=1.00\columnwidth]{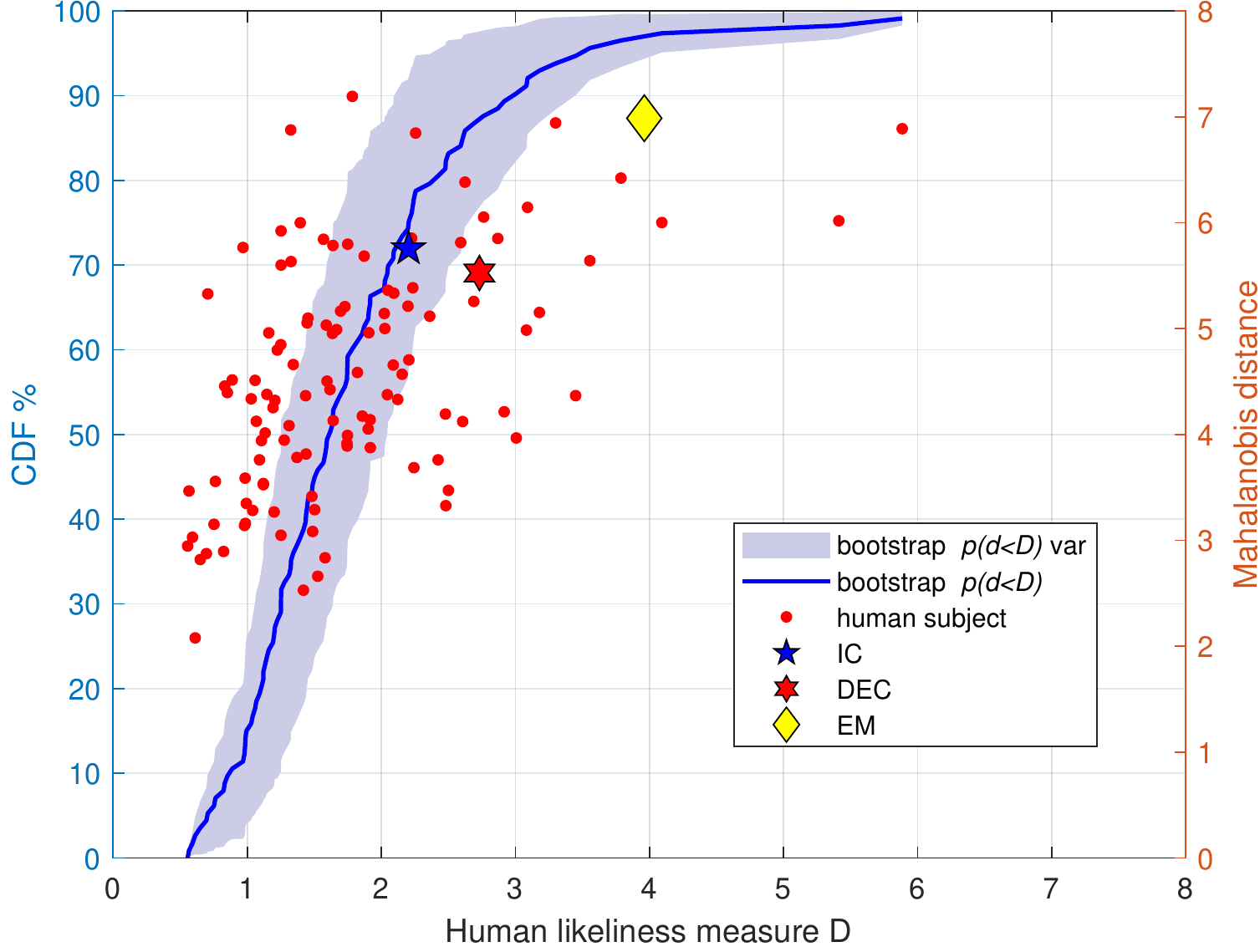}
	\caption{Mahalanobis distance for the samples in the data-set (red dots) as a function of the metric from eq. \ref{distance}, and the bootstrap estimated cumulative distribution of the metric (blue line) and its variance (light blue bands). The human-likeness score is smaller than the Mahalanobis distance because if the coefficients in Fig. \ref{fig:PRTSexample} D that are $\leq1$, and in general there is a spread of the Mahalanobis distances of samples associated with the same D because the weighting almost removes the variation due to high-frequency components. The three examples are shown for comparison.}
	\label{fig:dist}
\end{figure}

\textbf{The main contribution of the present work} is a set of data from human experiments that can be used as a benchmark for human-like posture control to be compared with robots, and the evaluation of a metric proposed for such comparison.
\\ \textbf{The tests for the bio-inspired control systems} showed that the three robot control systems produced small gains compared to humans: according to the proposed measure they were ``less human-like'' than most of the samples in the data-set. Notice that, as explained in the methods section, the parameters of the models can be optimized to create a FRF that is more similar to the human average (as it is actually done in several posture control studies that include a control model). Here, the aim of the simulation was not to evaluate the three models in general (the conclusion here is not that the IC is more human-like from an absolute point of view) or to have a ``fair competition'' between the three control concepts: that would require to perform the test on the real robot or, at least, optimize the control parameters based on the controlled system. The main idea was to get real models as they were used in real robot experiments (i.e. with parameters calibrated for that specific scenario) to test the proposed metric of human likeness.   
\\ \textbf{Human like behavior} can represent a standard for humanoid posture control, and with the data-set presented in this work, we propose a way to quantify it. Of course, this does not exclude the possibility that a robot can benefit from super-human capabilities or peculiar behaviors that are optimized for its non-human hardware, as motors are obviously different from muscles and almost all humanoid robots have limb kinematics that is simpler than the corresponding human one. Comparing FRFs is not the only method available in the literature. Models can be tested versus human data with statistical tests such as analysis of variance, focusing on some specific kinematic variables, as presented, for example, in \cite{bayon2020can} where the range of motions of the body links are considered. The direct comparison of FRFs has the main advantage of being based just on experimental data. In any case, mimicking human behavior may be intrinsically valuable in human-robot interactions for making movements understandable and physical interaction natural \cite{villalobos2017study}. The presented comparison principle can be extended using several FRFs obtained with different modalities, e.g. support surface translation and tilt, and different amplitudes. Performing different tests with the same control system parameters can highlight some specificity of human behavior such as nonlinear responses and sensory re-weighting (the importance of specific sensory inputs can change depending on the conditions). From the practical point of view, humans can get an advantage from a nonlinear response in the sense that smaller disturbances can be compensated in a more ``relaxed'' way saving energy and effort of the nervous system. This means that the stimulus-to-sway gain may be relatively larger for smaller stimuli, and smaller for larger and more dangerous perturbations.
\\ \textbf{Human likeness is a ``human likelihood''} once it is defined in terms of the distance between the tested sample and a distribution of given human behaviors. In particular the score in eq. \ref{distance} resembles a Mahalanobis distance $D=\sqrt{\Delta^T \Sigma^{-1} \Delta}$, but with the addition of weights. Assuming a joint normal distribution for the weighted FRFs, $\mathcal{N}(\mu,\Sigma)$, the Mahalanobis distance defines probability density, where a smaller distance associated with higher probability density%
\ifdefined\review
. 
\else
\footnote{Normality of variables describing posturography may be implicitly assumed when the effects on posture are tested through an analysis of variance, e.g. \cite{akccay2021visual,jilk2014contribution}.}.
\fi 
It comes naturally to ask how the proposed metric is distributed over the data-set and how it relates to the likelihood of a sample. Fig \ref{fig:dist} shows the distribution of the human likeness metric over the data-set of human experiments (cumulative distribution function and relationship between the metric and the Mahalanobis distance) together with the results on the tested control systems. 

The metric we used in this work opens the possibility to make geometrical considerations for the performance score: i.e. in terms of ``How much two behaviors are different?'' or, ``Which is the smallest change in the FRF that would bring the behavior to a certain score?''. This represents an advantage when one aims to define solutions with better scores. 
\\ \textbf{Translation and tilt} of the body support surface can be used in combination. Exploiting the structure of the PRTS described in \S \ref{dataset}, two PRTS signals can be designed with no overlapping peaks $P(f_{Peaks})$ in the power-spectrum. These two signals can be used together as profiles for the two stimuli and hence two independent FRF can be produced in a trial. Preliminary studies suggest that, although human responses to perturbations are in general non-linear, they may exhibit a linear superposition effect between these two modalities.
\\ \textbf{Future work} will focus on tests with real humanoid robots as soon as possible.  In this work, simulations were used in lieu of real experiments because of the limited laboratory access due to the pandemic. A performance indicator evaluating transient response will be developed using a raised cosine velocity profile instead of the PRTS (that is oriented to the study of steady-state behavior). This kind of stimulus represents a ``softer'' and safer version of a step function input and it was used in previous experiments with human subjects \cite{lippi2014modeling}.

\section{\uppercase{Data and Code}}
This performance indicator and the data-set used to define the standard for the benchmark are available
\ifdefined\review
\textit{Links have been removed to comply to the requirement of double-blind review}
\else
through the EUROBENCH project {http://eurobench2020.eu/contact-us/}. The EUROBENCH framework includes several tests and performance indicators provided with the aim of testing the performance of robots at any stage of development.
\fi
\section*{\uppercase{Acknowledgment}}
\ifdefined\review
\textit{Acknowledgments have been removed to comply to the requirement of double-blind review}
\else
 This work is supported by the project 
\begin{wrapfigure}{l}{0.08\columnwidth}
		{\includegraphics[width=0.12\columnwidth]{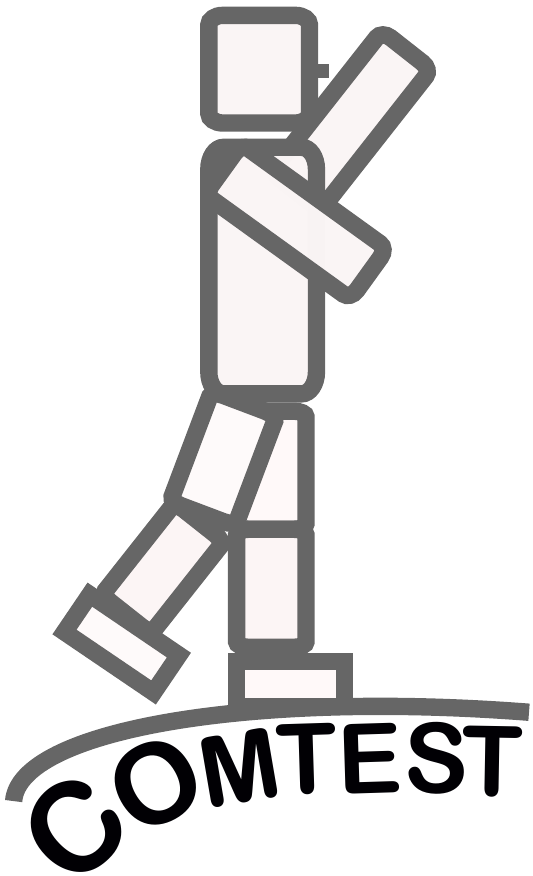}}
	\label{LOGO}
\end{wrapfigure}
\noindent COMTEST, a sub-project of EUROBENCH (European Robotic Framework for Bipedal Locomotion Benchmarking, www.eurobench2020.eu) funded by H2020 Topic ICT 27-2017 under grant agreement number 779963.
\fi

\bibliographystyle{apalike}
{\small
\bibliography{example}}

\end{document}